# Referring Expression Comprehension: A Survey of Methods and Datasets

Yanyuan Qiao, Chaorui Deng, Qi Wu*

*Abstract*—Referring expression comprehension (REC) aims to localize a target object in an image described by a referring expression phrased in natural language. Different from the object detection task that queried object labels have been pre-defined, the REC problem only can observe the queries during the test. It is more challenging than a conventional computer vision problem. This task has attracted a lot of attention from both computer vision and natural language processing community, and several lines of work have been proposed, from CNN-RNN model, modular network to complex graph-based model. In this survey, we first examine the state-of-the-art by comparing modern approaches to the problem. We classify methods by their mechanism to encode the visual and textual modalities. In particular, we examine the common approach of joint embedding images and expressions to a common feature space. We also discuss modular architectures and graph-based models that interface with structured graph representation. In the second part of this survey, we review the datasets available for training and evaluating REC approaches. We then group results according to the datasets, backbone models, settings so that they can be fairly compared. Finally, we discuss promising future directions for this field, in particular the compositional referring expression comprehension that requires more reasoning steps to address.

## I. INTRODUCTION

LANGUAGE and vision are closely related in daily life, and the referring expressions are often used in our social conversation and professional interaction, such as "please give me the white cup on the table". Referring expressions describe relationships between multiple objects in an image via natural language expressions, combining computer vision and computational linguistics.

While great progress has been made in bridging computer vision and natural language processing, the task of referring expression comprehension remains challenging because it requires a comprehensive understanding of complex language semantics and various types of visual information, such as the relationships between objects, attributes, and regions. In referring expression comprehension task, objects in an image are usually queried based on their category, attribute and context. Natural language based referring expressions can encode rich information, such as relationships that distinguish object instances from each other. More importantly, when the image contains multiple instances of the same object category,

Y. Qiao, C. Deng and Q. Wu are with the Australian Institute for Machine Learning, School of Computer Science, The University of Adelaide, Adelaide, SA 5005, Australia (e-mail:yanyuan.qiao@adelaide.edu.au, chaorui.deng@adelaide.edu.au, qi.wu01@adelaide.edu.au).
* corresponding author

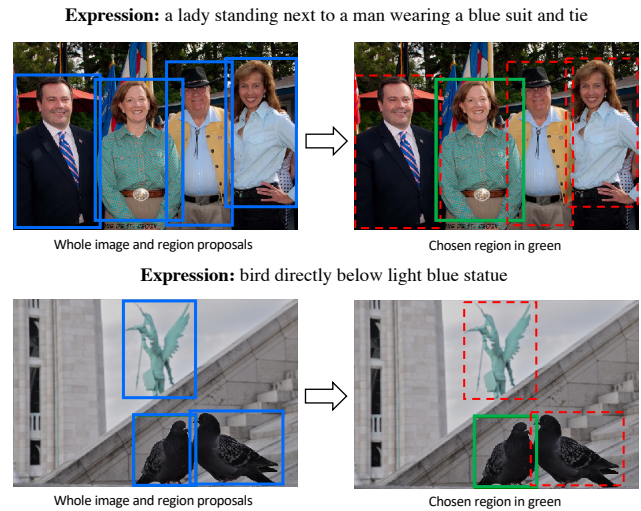

Fig. 1. Illustration of the Referring Expression Comprehension (REC) task. Given a referring expression and an image, REC aims to localize the referential entity. Blue boxes are candidates and green boxes are ground truth targets.

the referring expression is asked to be unambiguous so that it can distinguish the referenced instance from other instances. Referring expression is normally associated with three tasks: generation, segmentation and comprehension.

Referring expression generation (REG) aims at generating a discriminative description of an object in an image, which is very similar to the image captioning task.

Different from general image captions [1], [2], [3], [4], [5], [6], referring expressions are more specific about an object or region in the image. In addition, the generic image captioning lacks standard evaluation criteria, while the REG has a well-defined evaluation metric: it should describe an object unambiguously so that the system could easily comprehend the description, and infer and localize the right region in image. On the other hand, Referring expression segmentation (RES) aims to segment the referenced objects according to the referring expression [7]. The traditional solution is to embed the language encoder into the segmentation network, and then learn a multi-model tensor to decode the segmentation mask.

Referring expression comprehension (REC) is the reverse task of REG, which aims at localizing objects in an image based on referring expressions. As shown in Figure 1, REC is typically formulated as selecting the best region from a set of region proposals extracted from the image based on a referring expression. It is a challenging task considering the following aspects. First, the length of the expression can be as short as several words or can be a multi-round conversation. When the length of the expression is longer, the amount of information needed to be processed will also become larger and the system may require more reasoning steps to analyze it. Second, the image is high dimensional, and generally noisier than pure text. In addition, images lack the structure and grammatical rules as language. Thus, there are no effective and convenient analysis tools compared to those in NLP tasks such as syntactic parsers.

There are also some tasks similar to REC, such as visual grounding and object detection [8], [9]. Visual grounding is to localize multiple object regions in an image corresponding to multiple noun phrases from a sentence that describes the underlying scene. While the goal of referring expression comprehension is to find the best matching region by the given expression. Unlike object detection that uses predefined category labels to classify fixed objects, REC uses natural language expressions to refer to objects. This natural language expression is more practical because it varies according to the content of images and texts (such as categories, attributes, spatial configuration, and interaction with other objects), so it is more suitable for real application scenarios. The key of REC is to use language information to distinguish target objects from other objects, especially objects of the same category.

For other vision and language tasks, such as Visual Question Answering [10], [11] and Visual dialogue [12], [13], referring expression comprehension is of great importance. Though they have diverse model architectures, the first step of these tasks is to locate the object corresponding to the language description or question. Since the textual information is not a separate label, a simple detection method cannot meet the requirements. It requires the model to fully understand the natural language expression, including object and relation, which is an important step. Therefore, the matching between the expression and object is crucial to the above task. In this survey, we mainly focus on the referring expression comprehension task.

The increasing maturity of computer vision and NLP technologies, and the emergence of related large-scale datasets, have driven the growing interest in the community of REC research. As a result, there has been a great number of literature on REC in the past few years. The purpose of this survey is to give a comprehensive overview of the field, including methods, datasets, and suggestions of future directions. To the best of our knowledge, this article is the first survey in the field of REC. In Section II of this survey, we conduct a comprehensive review of methods of REC. These methods are divided into seven categories: joint embedding approaches, modular-based approaches, graph-based approaches, approaches using external parsers, weakly supervised approaches, one stage approaches, vision-language pre-training approaches. We introduce the above approaches, including their motivations, technical details, experimental results and limitations, respectively. In Section III, we describe currently available REC datasets and the evaluation metrics used for referring expression. In Section IV, we discuss possible future directions in REC.

## II. METHODS FOR REFERRING EXPRESSION COMPREHENSION

We summarize existing approaches to referring expression in Figure 2.

### A. Joint embedding approaches

#### 1) CNN/LSTM framework

**Motivation** Intuitively, to perform REC, the first step is to encode the image regions and the referring expression into the same vector space. For the representation of images, Convolutional Neural Networks (CNN) [45] can generate rich image representations by embedding input images into fixed-length vectors, so that these representations can be used for various visual tasks. For text representation, long short-term memory (LSTM) [46] networks have been widely used in sentence feature encoding and have shown promising performance on many sequential modeling tasks.

**Methods** The conventional way to referring expression is to use the CNN-LSTM framework. In the generation task, a pre-trained CNN network is used to model the target object and its context in the image, and then feed this as input to a LSTM for generation. In the comprehension task, LSTM takes a region-level CNN feature and a word vector as input, and at each time step, its purpose is to maximize the likelihood of the expression for a given referred region (see Figure 3).

Mao et al. [18] introduce the first deep learning approach to referring expression generation and comprehension. In this model, the authors utilize a CNN model to extract visual features from a bounding box and LSTM network to generate expressions. The CNN model is pre-trained on ImageNet [47], and the image features are from the entire image containing the target object. In the baseline method, the model utilizes maximum likelihood training, in which the probability of referring expression is only high for the reference region and low for all other regions. They also propose a maximum mutual information (MMI) method, if the generated reference expression can also be generated by other objects in the image, the model will be penalized. In addition, Mao's work also solves the inverse problem of referring expression comprehension, which aims at localizing the referred object in the image based on expression.

Similar to Mao's method, Yu et al. [19] also tackle the problem of referring expression generation and comprehension together. The difference is that they introduce a visual comparative method (Visdif) to distinguish the target object from the surrounding objects, rather than extract CNN features for the entire image. Visdif focuses on the relationship between the type of objects and expressions, and uses this visual difference between objects to represent the visual context to help improve performance. These visual comparisons are essential for generating unambiguous expressions, since visual characteristics of similar objects must be considered in the generation process in order to select the most different aspects to describe.

Nagaraja et al. [20] extend Mao's work to learn context regions via multiple-instance learning (MIL) since annotations for context objects are generally not available for training.

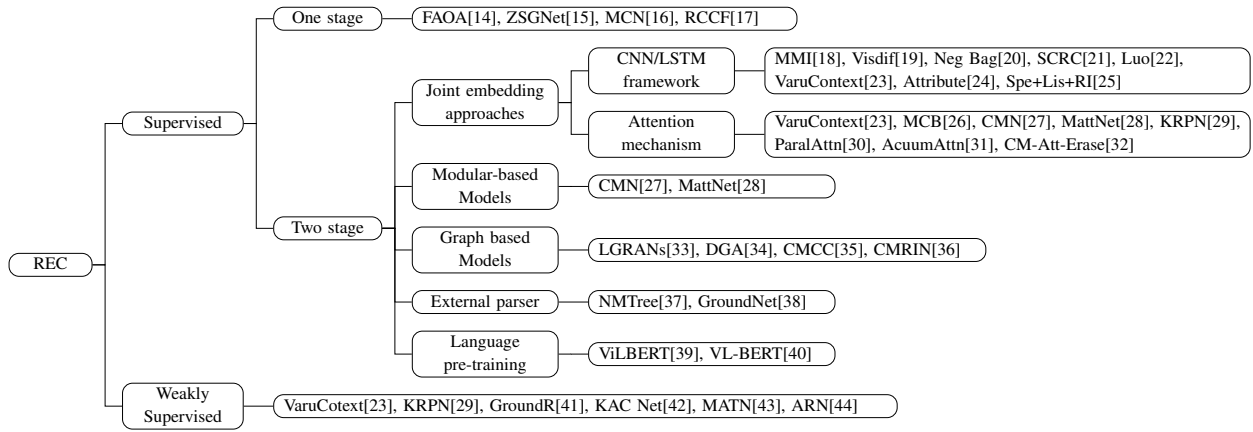

Fig. 2. Overview of existing approaches of Referring Expression Comprehension. Methods are categorised to different groups.

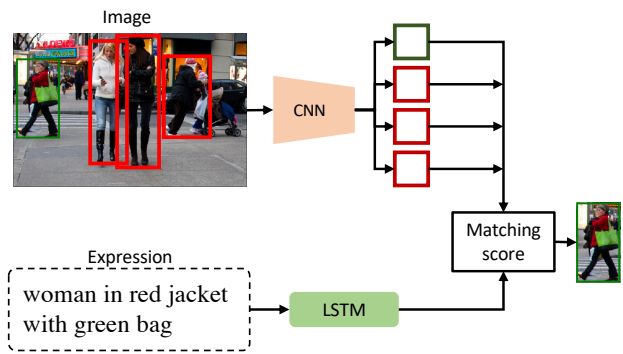

Fig. 3. A common approach to Referring Expression Comprehension (REC) using CNN-LSTM framework (Section II-A1).

They use the MIL object functions to train the LSTM. Different with Mao's method, the entire image is also regarded as a potential context region. Thus they input the word embedding and region features, as well as CNN features of the entire image and bounding box features as context. They use LSTM to model the probability of a referring expression as a function of a region and a context region, mapping the referring expression to a region and its supporting context region. The formulas for these two training objective functions are similar to MISVM and mi-SVM [48].

Hu et al. [21] propose a new Spatial Context Recurrent ConvNet (SCRC) model. This model takes natural language query, local image descriptors, spatial configurations and global scene-level context as input to score the output of candidate regions. Inspired by the long-term recursive convolutional network (LRCN), SCRC uses a two-layer LSTM network structure in which the embedded text sequence and visual features are used as the input of the first and second layers. Due to the lack of annotated object bounding boxes and description pairs in the dataset, SCRC pre-processes the image caption domain, and then adapts it to the natural language object retrieval domain, so that the visual language knowledge can be transferred from the previous task can be transferred in the latter task. This pre-training and adaptive process improves performance and avoids over-fitting.

Luo et al. [22] utilize learned comprehension models to guide the model generate better referring expressions. The comprehension module trained on human-generated expressions, and trained by proxy method. The training by proxy method is inspired by the generator-discriminator structure in Generative Adversarial Networks [49], [50]. Then they use a generate-and-rerank pipeline, in which the comprehension model serves as a "critic" to tell if the expression can be correctly reference. To be specific, the generate-and-rerank pipeline first generate candidate expressions, and then select the best one according to the performance on the comprehension task.

Zhang et al. [23] propose a variational Bayesian method named Variational Context (VC), utilizes the relationship between the referent and context. Specifically, either referent or context will affect the estimation of the posterior distribution of the other, thus the search space of the context will be reduced. For each region, VC estimate a coarse context, which will help to refine the true localization of the referent. The variational lower bound can be interpreted by the interaction between the reference and the context: given any of them can help localize the other, so it is expected to reduce the complexity of the context. Compared with formal MIL-based methods, VC uses this framework to approximates the combinatorial context configurations with weak supervision.

**Performance and limitations** We summarize performances of all discussed methods and datasets in Tables II-VII. As one of the earliest introduced methods, Mao's work is considered the *de facto* baseline result. It works well from two-word phrases to longer descriptions and can be trained in a semi-supervised manner by automatically generating descriptions of image areas. The "VisDif" trains a Fast-RCNN detector and conduct experiments based on it. The results show that this combination of related objects produces better results than previous works to reduce ambiguity during generation. One weakness is that the model relies on detection performance. Nagaraja's work shows that context modeling between objects provides better performance than just modeling object attributes. Their model could identify a referring region and its supporting context region. "SCRC" method effectively uses local and global information, significantly improve performance than previous baseline methods on ReferItGame dataset and Flickr30K Entities dataset. The results show that the incorporation of spatial configuration and global context can improve the performance. Luo's work improves referring expression generation on multiple benchmark datasets. "VaruContext" also extend the model to unsupervised settings, the performance improves on ReferItGame. A large number of experiments on a variety of benchmarks show continuous improvement over past methods in both supervised and unsupervised settings.

The CNN-LSTM framework is concise and effective. However, these approaches are limited by the monolithic vector representations that ignore the complex structures in the compound language expression as well as in the image. That is to say that for an expression, they encode the expression sequentially and ignore the dependencies in the expression.

*2) Attention mechanism*

**Motivation** Attention mechanisms have been applied on many vision-language applications [51], [52]. It allows machine to focus on a certain part of input when processing a large amount of information. After being introduced into the Referring Expression Comprehension field, the attention mechanism has brought out many breakthroughs [16]. Specifically, it is able to build element-wise connections between visual and textual information, so that one can utilize the information from some specific image regions (*i.e.*, regions of interests) when encoding each word in the text and vice versa, leading to semantically-enriched visual and textual representations.

**Methods** Zhuang et al. [30] propose a Parallel Attention (PLAN) approach that contains two parallel attention process reasoning at both image-level and region-level. Given a referring expression, the image-level attention module encodes the entire global information and the referring language descriptions by recurrently attending on different image patches, which allows the model to utilize the relevant contextual information. While the region-level attention module recurrently attends on the object proposals based on the language information, and obtain a representation for each proposal. Then, the representations of the global image and the object proposals are put into a matching module to compute a matching probability for each object proposal.

Deng et al. [31] formulate REG into three sequential sub-tasks, *i.e.*, 1) identify the main focus in the query; 2) understand the concepts in the image; 3) locate the most relevant object, and they proposed an Accumulated Attention (AccumAtt) mechanism to solve these three tasks jointly. Specifically, AccumAtt adopts three attention modules to handle the attention problem in the above three sub-tasks, respectively. To capture the correlations among those attention problems, AccumAtt further employs an accumulating process to combine all types of attention together and refine them circularly, where each type of attention will be utilized as a guidance when computing the other two. Finally, AccumAtt computes a matching score between the attended representation of each object proposal and the refined representations from the query and image attention modules to locate the target object.

Attention mechanisms can be also combined with other

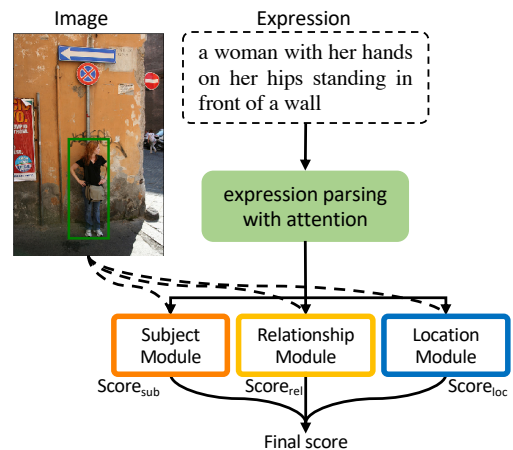

Fig. 4. The Modular-based model (Section II-B) parses expression into three phrase embeddings, such as subject, relationship, location or object embeddings. Then these embeddings are input to three modules that process the visual region in different ways. After compute individual matching scores, the model combines the module scores to output the final score.

models such modular networks, graph models, which will be introduced in the following sections.

**Performance and limitations** By applying attention mechanism on top of the CNN/LSTM framework, AccumAtt [31] and PLAN [30] both achieve remarkable performance improvements on RefCOCO, RefCOCO+, RefCOCOg and GuessWhat?! dataset. Besides, the experiments in PLAN verifies the importance of performing attention at multiple levels, *i.e.*, region-level and image-level. Also, the ablation studies of AccumAtt shows the effectiveness of using multiple attention modules to handle different modalities and accumulating the attention scores for multiple rounds. Moreover, the visualization results of these two approaches also provide valuable interpretations for understanding the attention process.

However, these attention-based methods cannot guarantee a correct attention assignment since the dataset usually provides no corresponding annotations.

*B. Modular-based Models*

**Motivation** Modular networks have been successfully applied to many tasks such as visual question answering [53], visual reasoning [54], relationship modeling [27], and multi-task reinforcement learning [55]. For example, for VQA tasks, the modular networks decompose the question into several components and dynamically assemble a network to compute an answer to the given question, which shows strengths than competing approaches. In the task of referring expression, previous works mainly simply concatenate all features as input, and use a single LSTM to encode or decode the whole expression, ignoring the variance among different information provided in expressions. While the early work requires an external language parser [56] for decomposition, recent methods suggest learning end-to-end decomposition. By decomposing the expression into different components and matching each component with the corresponding visual region through a modular network, one-step reasoning is achieved (see Figure 4).

**Methods** Hu et al. [27] propose Compositional Modular Networks (CMNs) which learns linguistic analysis and visual inference end-to-end. CMNs are composed of a localization module and a relationship module. The localization module matches subject or object with each image region and returns a unary score. And the relationship module matches a relationship with a pair of regions and returns a pairwise score. Given an image and an expression, CMNs firstly parses the expression into a subject, relationship and object with soft attention. Then these two modules are combined to consider local characteristics of regions and interactions between regions.

Yu et al. [28] propose the Modular Attention Network (MAttNet). MAttNet decomposes natural language expressions into three modular components, which are related to subject appearance, location and relationship to other objects. The subject module deals with category, colors, and other attributes, the location module handles absolute and relative locations, and the relationship module deals with subject-object relationships. Each module has a different structure and learns parameters in its own module space without affecting other modules. Instead of relying on an external language parser, MAttNet learns to automatically parse expressions through a soft attention based mechanism. Then the matching scores of the three visual modules are calculated to measure the compatibility between objects and expressions.

Liu et al. [32] propose a Cross-Modal Attention-guided Erasing (CM-Att-Erase) strategy for training referring expression comprehension models. CM-Att-Erase adopts the MAttNet as its backbone model. By erasing the most attended part from either textual or visual information during training, CM-Att-Erase encourage the model to discover more latent cues for visual-textual alignment, leading to a comprehensive cross-modal correspondence. Moreover, CM-Att-Erase makes several modifications to the design of each module. Specifically, unlike MAttNet which only adopts the textual information to learn word-level attention and module-level attention, CM-Att-Erase further considers the global image feature. Moreover, CM-Att-Erase formulates the location and relationship modules into a unified structure with sentence-aware attention, which considers multiple context objects, and attends to the most important ones.

**Performance and limitations** Experiments on the RefCOCOg dataset show that although weakly supervised, CMN not only can locate the subject region correctly, but also finds reasonable regions for the object entity. However, CMN relies on limited syntactic information, and due to the lack of modeling of language recursion, it is impossible to trace multiple supported objects. MAttNet has a remarkable superior performance, achieving about 10% improvements on the localization of the bounding box, and the precision pixel segmentation is almost doubled. MAttNet has become one of the most important baseline and backbone model in referring expression comprehension. CM-Att-Erase obtains significant improvements on top of MAttNet on RefCOCO, RefCOCO+ and RefCOCOg dataset, with both ground-truth object proposals and detected object proposals. Moreover, it still outperforms MAttNet by a clear margin without the cross-modal erasing training strategy.

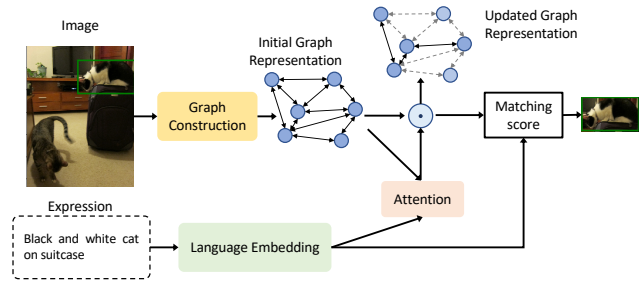

Fig. 5. Overview of the Graph based model for referring expression comprehension (Section II-C). The graph construction module builds a graph over the candidate objects, where the nodes and edges correspond to the objects and relationships. And then the language representation of the expression is fused into the graph to obtain the updated graph representation. Finally, the matching module computes the matching score between objects and expression.

The CM-Att-Erase proposed an attention based erase process to discard the most dominant information from either textual or visual domains to generate difficult training samples online, and to drive the model to discover complementary textual-visual correspondences. This process encourages the model to explore harder examples, eliminating the dataset bias. In addition, the modular-based models, such as CM-Att-Erase achieve better results than more sophisticated approaches, such as graph-based models. This suggests that the system can achieve a good performance without complex reasoning process under the condition of limited existing datasets.

The limitation of the method is that these modular networks oversimplify the language structure and ignore the relationship between visual objects. In other words, the language and regional features are independently learned or designed without knowing each other, which makes it difficult for the features of the two patterns to adapt to each other, especially in the case of complex expressions. Therefore, this method is not applicable when the expression is over complex.

*C. Graph based models*

**Motivation** For the referring expression task, the key to solving this problem is to learn a distinguishing object feature that can adapt to the expression used. To avoid ambiguity, this expression usually describes not only the properties of the object itself, but also the relationship between object and its neighbors. Previous methods deal with objects alone, or only study first-order relationships between objects, without considering the potential complexity of expressions. Therefore graph-based methods are proposed, in which the nodes can highlight related objects, and the edges are used to recognize the object relationship existing in the expression. Because graph attention can build references and other supporting cues on the graph, reference expression decisions can be both visual and interpretable (see Figure 5).

**Methods** Wang et al. [33] propose a language-guided graph attention network (LGRAN). The network consists of three modules: language-self attention module, language-guided graph attention module, and matching module. The language self-attention module adopts a self-attention scheme to decompose the expressions of three parts describing the subject, intra-class relationships and inter-class relationships, and learn the corresponding representations. The language-guided graph attention module, constructing a candidate objects of a directed graph, emphasizing the nodes, intra-class edges and inter-class edges, and finally get three types of expression-relevant representations for each object. The matching module calculates the expression-to-object matching score. LGRAN can dynamically enrich the object representation on the basis of attended graph to better adapt to the referring expressions.

Yang et al. [34] propose a Dynamic Graph Attention Network (DGA) which enable the multi-step reasoning of the interactions between the image and expression. Given an expression and image, DGA builds a graph of the objects in the image, where the nodes and edges correspond to the objects and relationships, then merges the language representation of the expression into the graph. After that, the analyzer learns the language guidance of reasoning by exploring the language structure of the expression. Under the guidance of the predicted visual reasoning process, DGA performs dynamic inference step by step on the graph. The visual reasoning process is a sequence composed of constituent expressions. Finally, DGA calculates the matching score between the compound object and the referring expression.

Similarly, Liu et al. [35] utilize a language-guided graph representation to capture the global context and its interrelationships of grounding objects. In addition, they also propose a context-aware cross-modal graph matching strategy named CMCC. CMCC consists of four main modules: a backbone network extracts basic language and visual features, a phrase graph network encodes phrases in sentence descriptions, a visual object graph network calculates object suggestion features, and a graph similarity network predicts the global matching between phrases and object suggestions.

Yang et al. [36] propose a Cross-Modal Relationship Inference Network (CMRIN) which contains a Cross-Modal Relationship Extractor (CMRE) and a Gated Graph Convolutional Network (GGCN). CMRE adaptively extracts all the required information to build a language-guided visual graph with cross-modal attention. GGCN fuses information from different modes and propagate multi-modal information to calculate semantic contexts.

**Performance and limitations** LGRANs makes reference expression decision making both visual and interpretable. Experiments have been performed on the datasets (RefCOCO, RefCOCO+, and RefCOCOg), and the results have certain advantages. The DGA experimental results show that this method can not only significantly surpass all existing algorithms, but also provide visual evidence for gradually locating objects in complex language descriptions. CMCC trains the entire graph neural network in a two-stage strategy and evaluates it on the Flickr30K entity benchmark, achieving the state-of-the-art performance and surpassing other methods with a large advantage.

### D. External parser

**Motivation** Referring expression uses the discrimination characteristics of the target object and its relative position with other objects to disambiguate the target object. There is a correspondence between the linguistic structure of the referential expression and the parsing process of the referential expression. Unfortunately, existing methods oversimplify the compound nature of language, reducing it to a single sentence, or composite scores for subject, predicate, and object phrases. Although some of them use vocabulary-level attention mechanisms to focus on the information language part, their reasoning is still rough compared to human-level reasoning. More seriously, this rough basic score is easily biased towards learning a certain visual language model rather than visual reasoning. This problem has been repeatedly found in other tasks used in many end-to-end visual language embedding frameworks, such as VQA and image captioning. Therefore, with the aid of external language parser, the grammar-based methods can help locate the target object and auxiliary supporting objects mentioned in the expression (see Figure 6).

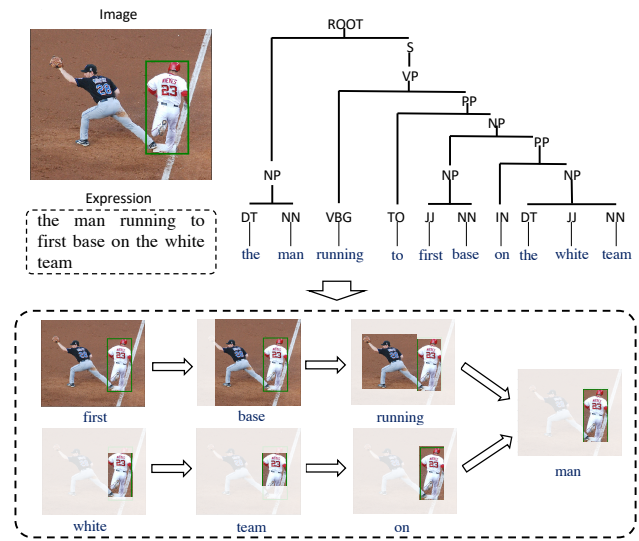

Fig. 6. Overview of the model with external parser for referring expression comprehension (Section II-D). The model regularizes the REC task along the dependency parsing tree of the sentence, and computes visual attention according to its linguistic feature. Then the model localizes the target object by accumulating the grounding confidence score along the dependency parsing tree in a bottom-up direction.

**Methods** Cirik et al. [38] introduce GroundNet, which is the first approach that uses external syntactic analysis to parse input reference expressions, which aids the localization of target objects and auxiliary support objects. When given a parse tree of input expressions, they explicitly map the syntactic components and relationships that appear in the tree to a composed graph composed of neural modules, which defines the architecture for performing localization.

Liu et al. [37] develop a Neural Module Tree network (NMTree), which accumulates the grounding confidence in the Dependency Parsing Tree (DPT) [57] of natural language sentences to locate the target region. Given a natural language expression as input, they first transform it into NMTree by Dependency Parsing Tree, Bidirectional Tree LSTM, and Module Assembler. NMTree consists of three simple neural modules: Single for the leaves and root, and the internal nodes are Sum and Comp. Each module calculates a grounding score and accumulates it in a bottom-up fashion. The final result grounding score is the output score of the root node.

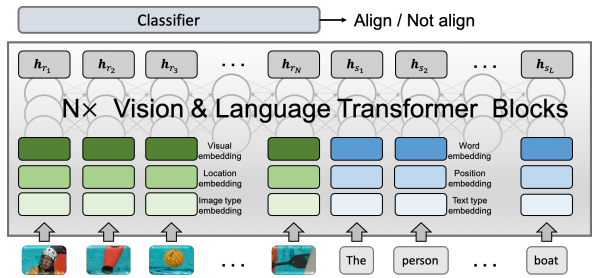

Fig. 7. A typical Vision-Language Pre-training framework for Referring Expression Comprehension (Section II-E). The essential of the model architecture is a stack of transformer-like modules that can model multiple modality jointly. In this example, the features of candidate regions as well as the word embeddings of the text tokens are fed together into the vision-language transformer model to capture the relations among the image regions and text tokens. Then, a classifier is applied on top of the hidden state of each candidate region to predict its matching score.

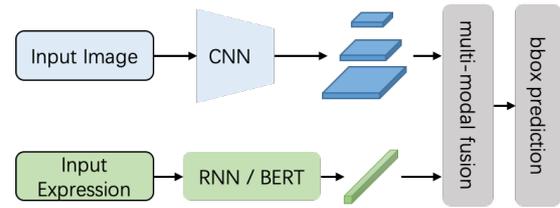

Fig. 8. General framework of One-stage Referring Expression Comprehension (Section II-F). The pixel-wise image feature and text feature are fused together to obtain a heatmap for the region-of-interest. Afterwards, a target proposal module is adopted to predict the target bounding box from the heatmap.

**Performance and limitations** Compared with previous methods, GroundNet has better interpretability. First, it can determine which phrase of the referring expression point points to which object in the image, and secondly it can track how the location of the target object is determined by the network. GroundNet maintains comparable performance in the localization of target objects. Qualitative results and manual evaluations show that NMTree understands more fine-grained and interpretable language compound reasoning and has better performance. However, when using a parser to process and transform an expression, some determiners may be removed, while these vocabularies such as the indefiniteness of a noun may help determine the object.

### E. Vision-Language pre-training

**Motivation** Vision-language reasoning requires the understanding of visual contents, language semantics, and cross-modal alignments and relationships. The dominant strategy in this area is to utilize separate vision and language models that are specifically designed for vision or language tasks and pre-trained on large-scale vision or language datasets, which however, lacks a unified foundation for learning joint representations among visual concepts and language semantics, often resulting in poor generalization ability when paired vision-language data is limited or biased. Thus, it could be beneficial to have a jointly pre-trained vision-and-language model to provide a unified knowledge representation for downstream vision-and-language tasks (see Figure 7).

**Methods** Lu et al. [39] propose the ViLBERT framework to perform joint modeling for vision and language information on a large-scale vision-language dataset and leverage the pre-trained model to solve multiple vision-language tasks including referring expression comprehension. Specifically, ViLBERT contains two branches, one for language information and the other for vision information. Each stream consists of two different layers: the Transformer Layer (TRM) and the proposed Co-Attentional Transformer Layer (Co-TRM), where the latter is used for connecting the two branches. To learn joint vision-language representations, ViLBERT first obtain a set of object proposals from the input image with a pre-trained object detector. The Co-TRM layers then calculate cross-modality attention scores between each object proposal and each token in the input text. The pre-training is performed on Conceptual Captioning, a large-scale image captioning dataset that contains 3 million image-captioning pairs. Two pre-train tasks are used: 1) Masked Multi-modal Learning, where the model must reconstruct image region categories or words for masked inputs given the observed inputs; 2) Multi-modal alignment prediction, where the model must predict whether or not the caption describes the image content. When finetuning on referring expression comprehension task, they simply pass the final representation for each image region into a learned linear layer to predict a matching score.

Different with ViLBERT that adopts separate branches for different input modalities and perform cross-modal attention and self-attention alternately to obtain the joint representations, Su et al. [40] propose to process the image and text input in an unified single-branch framework named VL-BERT. In VL-BERT, each object proposal in the image or each word in the text is treated as an input token into a stack of Transformer Layers, which learns to build connections among all types of tokens without any restrictions, thus the co-attention and the self-attention are performed simultaneously. Like in ViLBERT, the pre-training of VL-BERT is also performed on Conceptual Captioning, and VL-BERT only adopt the Masked Multi-modal Learning as the pre-training task. When finetuning on referring expression comprehension task, VL-BERT follows the same strategy as in ViLBERT.

**Performance and limitations** With vision-and-language pre-training, ViLBERT and VL-BERT achieve the best performance on RefCOCO+ dataset and outperform the previous state-of-the-art results by a significant margin. However, their models are generally very large and require a high computational budget in practice.

### F. One stage approaches

**Motivation** Most referring expression comprehension approaches follow a common two-stage pipeline: i) use an object detector like Faster RCNN to generate a set of object proposals based on the input image; ii) compute a matching score between each proposal and the referring expression, after which the proposal with the highest score is adopted as the model prediction. However, these two-stage solutions have two notable issues. First, the object detection phase introduces additional computations. Besides, since there are normally tens or even hundreds of proposals generated in the stage-i, huge among of computation is required to extract features for all the detected object proposals, which makes it infeasible for real-time referring expression comprehension. Second, the quality of the object proposals obtained in stage-i can severely affect the final performance. In fact, since the state-of-the-art object detectors still yield limited performance in practice, there usually exists lots of misalignment between the generated proposals and the ground-truth objects, which may impede the learning procedure in stage-ii. More critically, the object detector may even miss the target in stage-i (which is quite common), in which case the stage-ii is actually a wasting of time since the VG task will never succeed (see Figure 8).

**Methods** Yang et al. [14] develop a one-stage referring expression comprehension approach (FAOA) based on YOLOv3 [58]. Specifically, they obtain a feature pyramid from the image through Darknet [58] and extract the language features from the referring expression through BERT. Then, for each level in the image feature pyramid, the language feature and the image feature are concatenated together and fed into the prediction head to predict the bounding box of the target object. To make the model have a sense of location, a spatial feature that consists of the normalized coordinates of the pixels is also fed into the prediction head.

Similarly, Sadhu et al. [15] propose a Zero-Shot Grounding (ZSG) method that combines the detection network and the grounding system together. Different with [14], they adopt a ResNet with FPN to extract the image feature pyramid and use a Bi-LSTM to extract the language representation. Moreover, instead of Binary Cross Entropy, they adopt Focal Loss to classify the regions inside/outside the target object.

Luo et al. [16] propose a Multi-task Collaborative Network (MCN) that jointly solves the referring expression comprehension and the referring expression segmentation in a one-stage pipeline. Specifically, they seek to address the prediction conflict problem between the two referring expression tasks through 1) Consistency Energy Maximization, which forces the two tasks to have similar attentions on the input image; 2) Adaptive Soft Non-Located Suppression, which softly suppresses the response of unrelated regions in referring expression segmentation branch according to the predictions in referring expression comprehension branch.

Liao et al. [17] propose a Realtime Cross-modality Correlation Filtering method (RCCF) that reformulate the referring expression comprehension task as a correlation filtering process. RCCF compute a correlation map between the referent and the image to predict the object center.

**Performance and limitations** Compared with two-stage referring expression comprehension methods, one-stage methods have significantly faster inference speed. In fact, the most time-consuming part in two-stage approaches is the generation of the object proposals and the feature extraction of them, while in one-stages approaches these two steps are not required, leading to an order of magnitude speedup. Moreover, FAOA and MCN achieves competitive or even better performance with several strong two-stage baselines including MAttNet, the previous state-of-the-art model.

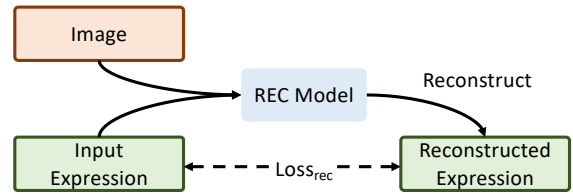

Fig. 9. Framework of the Weakly supervised model (Section II-G). The model trains the system by learning to reconstruct the language information contained in the input expression from the predicted object.

### G. Weakly supervised

**Motivation** Traditionally, training referring expression models in a supervised manner requires large numbers of manual annotations indicating the mapping between input expressions and objects. In addition, the supervised model can only handle certain types of grounding in limited training data, and cannot meet the needs of practical applications. Therefore, it is desirable to learn from data with no or little supervision. Weakly supervised referring expression refers to locating the object in the image according to the language query, when the mapping between the reference object and the query is unknown in the training phase (see Figure 9).

**Methods** Rohrbach et al. [41] propose GroundR that learns grounding by reconstructing a given phrase using an attention mechanism. During training, GroundR consists of two stages. The first stage aims to select the most relevant region or potentially multiple regions based on the phase, and then the second stage attempts to reconstruct the same phrase from regions it attended to in the first stage. At test time, GroundR removes the phrase reconstruction stage and only uses the first part for grounding. However, they ignore the discriminative information of location and context from the referential object, and cannot distinguish a specific object where multiple objects of a particular category situated together.

Based on GroundR, Chen et al. [42] design a Knowledge Aided Consistency Network (KAC Net), which reconstructs both the input expression and region's information. KAC Net consists of two branches: a visual consistency branch and a language consistency branch. Visual consistency branch aims at predicting and aligning expression-related region's location parameters, and language consistency branch aims at reconstructing input expression from regions. In order to utilize the complementary knowledge from the visual feature extractor, a knowledge based pooling gate is used to focus on the visual and linguistic reconstruction.

Different from selecting the optimal region from a set of candidate regions, Zhao et al. [43] propose a Multi-scale Anchored Transformer Network (MATN), which locate free-form text phrases in image without bounding box supervision information. MATN consists of a multi-scale correspondence network and an anchor constraint induced from a set of candidate regions. The model is trained by simultaneously minimizing contrast reconstruction errors between different phrases in a single image and a set of triple losses between multiple images with similar phrases.

Liu et al. [44] propose an adaptive reconstruction network

TABLE I
MAJOR DATASETS FOR REFERRING EXPRESSION AND THEIR MAIN CHARACTERISTICS.

| Dataset | Source of images | Number of images | Number of expressions | Number of objects | Avg. length words | Object categories | Object of the same type | Number of attributes | Number of relations | Number of candidates | Dictionary |
|---|---|---|---|---|---|---|---|---|---|---|---|
| ReferItGame[59] | Image CLEF | 19,894 | 130,525 | 96,654 | 3.61 | 238 | - | - | - | - | 8,800 |
| RefCOCO[19] | MSCOCO | 19,994 | 142,209 | 50,000 | 3.61 | 80 | 3.9 | - | - | 10.6 | 2,890 |
| RefCOCO+[19] | MSCOCO | 19,992 | 141,564 | 49,856 | 3.53 | 80 | 3.9 | - | - | - | - |
| RefCOCOg[18] | MSCOCO | 26,711 | 104,560 | 54,822 | 8.43 | 80 | 1.63 | - | - | 8.2 | 4,849 |
| Flickr 30k entities[60] | Flickr30K | 31,783 | 158,915 | 275,775 | - | 44,518 | - | - | - | - | 17,150 |
| GuessWhat?[61] | MSCOCO | 66,537 | 155,280 | 134,073 | - | - | - | - | - | - | 1,000 |
| CLEVR-Ref+[62] | CLEVR | 99,992 | 998,743 | 492,727 | 22.4 | 3 | - | 12 | 5 | - | - |
| Cops-Ref[63] | COCO/Flickr | 75,299 | 148,712 | 1,307,885 | 14.4 | 508 | - | 601 | 299 | 262.5 | 1,596 |
| Touchdown-SDR[64] | Google Street View | 25,575 | 9,326 | 99,788 | 26.97 | - | - | - | - | - | - |
| Refer360°[65] | SUN 360 | 2000 | 17,137 | 124,880 | 43.80 | - | - | - | - | - | - |
| REVERIE[66] | Matterport3D | 10,318 | 21,702 | 4,140 | 18 | 489 | - | - | - | - | 1,600 |

(ARN), which adopts the method of adaptive grounding and collaborative reconstruction to establish the corresponding relationship between image proposal and queries. It consists of feature encoding adaptive grounding and collaborative reconstruction. Specifically, they first extract subject, location, and context features to represent the proposals and queries. Then the adaptive grounding module is used to calculate the matching score between each suggestion and query through the hierarchical attention model. Finally, based on attention scores and proposal features, they use the collaborative loss of language reconstruction loss, adaptive reconstruction loss and attribute classification loss to reconstruct the input query.
**Performance and limitations** GroundR evaluates on Flickr 30k Entities and ReferItGame dataset. The results show that GroundR can handle a small portion of the annotated data available, and take advantage of unsupervised data, using the same amount of labeled data. Its performance on the Flickr 30k entities and the ReferItGame dataset is 4.5% and 10.6% higher than the most advanced technology, respectively. While the accuracy of KAC Net has increased by 9.78% and 5.13%, respectively. Extensive experiments demonstrate that the MATN outperforms state-of-the-art weakly supervised phrase localization methods by a significant margin. The adaptive mechanism of ARN helps the model alleviate the difference between different referring expressions. The qualitative results show that the method can better deal with the simultaneous existence of multiple objects of the same category.

## III. DATASETS AND EVALUATION

A number of datasets have been proposed specifically for research on referring expression. In the following sections, we will review the available datasets and describe how the datasets are created and discuss their limitations. Key characteristics are summarized in Table I.

*a) ReferItGame:* REG has been studied for many years in linguistics and natural language processing, but mainly focused on small or artificial datasets [68]. In 2014, Kazemzadeh et al. [59] introduced the first large-scale dataset ReferItGame, containing referring expression in real-world scenes. ReferItGame consists of 130,525 expressions, referring to 96,654 distinct objects, in 19,894 images. Its dataset construction is based on the ImageCLEF IAPR [69] image retrieval dataset, and SAIAPR TC-12 [70] expansion that includes segmentation of each image into regions. This dataset was collected in a two-player game, in which the first player writes a referring expression based on the specified target object. The second player is shown only the image and the referring expression, and asked to click on the correct location of described objects. If the click is in the target object region, both players receive game point and they swap for the next image. If the click is not correct, they remain in their current roles. However, the images in this dataset sometimes contain only one object of a given class, so the speakers tend to use short descriptions and focuses on context rather than objects.

TABLE II
RESULTS COMPARISON ON REFERITGAME (DET MEANS EDGEBOX PROPOSALS).

| Model | Categories | Test | Test(det) |
|---|---|---|---|
| SCRC[21] | CNN/LSTM framework | 72.74 | 17.93 |
| GroundR[41] | Weakly Supervised | - | 26.93 |
| MCB[26] | Joint embedding | - | 28.91 |
| Luo[22] | CNN/LSTM framework | - | 31.84 |
| CMN[27] | Modular-based Models | 81.52 | 28.33 |
| VaruContext[23] | CNN/LSTM framework | 82.43 | 31.13 |
| ZSGNet(cls)(vgg16)[15] | One stage approaches | 53.31 | - |
| ZSGNet(cls)(resnet50)[15] | One stage approaches | 58.63 | - |
| FAOA[14] | One stage approaches | - | 59.30 |
| DDPN(vgg16)[67] | Joint embedding | - | 60.3 |
| DDPN(resnet101)[67] | Joint embedding | - | 63.0 |
| RCCF[17] | One stage approaches | - | 63.7 |

TABLE III
RESULTS COMPARISON ON REFERITGAME UNDER WEAK SUPERVISION(ACCURACY (IOU > 0.5) OF PHRASE LOCALIZATION). THESE METHODS BELONG TO WEAKLY SUPERVISED APPROACH.

| Model | Accuracy(%) |
|---|---|
| GroundR(LC)($VGG_{cls}$)[41] | 10.70 |
| MATN[43] | 13.61 |
| VaruContext[23] | 14.11 |
| KAC Net($VGG_{cls}$)[42] | 15.83 |
| ARN[44] | 26.19 |
| KPRN[29] | 33.87 |

*b) RefCOCO & RefCOCO+:* These datasets were collected using ReferIt Game [59]. The authors further collected RefCOCO and RefCOCO+ datasets on MSCOCO [71] images. The RefCOCO and RefCOCO+ datasets each contain 50,000

TABLE IV
COMPARISON WITH STATE-OF-THE-ART METHODS ON REFCOCO, REFCOCO+ AND REFCOCOG WHEN GROUND-TRUTH BOUNDING BOXES ARE USED.

| Method | Visual features | Categories | RefCOCO | | | RefCOCO+ | | | RefCOCOg | | |
|---|---|---|---|---|---|---|---|---|---|---|---|
| | | | val | testA | testB | val | testA | testB | val* | val | test |
| MLE[18] | vgg16 | CNN/LSTM framework | - | 63.15 | 64.21 | - | 48.73 | 42.13 | 55.16 | - | - |
| MMI[18] | vgg16 | CNN/LSTM framework | - | 71.72 | 71.09 | - | 58.42 | 51.23 | 62.14 | - | - |
| Visdif[19] | vgg16 | CNN/LSTM framework | - | 67.57 | 71.19 | - | 52.44 | 47.51 | 59.25 | - | - |
| Visdif+MMI[19] | vgg16 | CNN/LSTM framework | - | 73.98 | 76.59 | - | 59.17 | 55.62 | 64.02 | - | - |
| Luo[22] | vgg16 | CNN/LSTM framework | - | 74.04 | 73.43 | - | 60.26 | 55.03 | - | 65.36 | - |
| Neg Bag[20] | vgg16 | CNN/LSTM framework | 76.9 | 75.6 | 78.0 | - | - | - | - | 68.4 | - |
| Attribute[24] | vgg16 | Attention mechanism | - | 78.85 | 78.07 | - | 61.47 | 57.22 | 69.83 | - | - |
| VaruContext[23] | vgg16 | Attention mechanism | - | 78.98 | 82.39 | - | 62.56 | 62.90 | 73.98 | - | - |
| AccumulateAttn[31] | vgg16 | Attention mechanism | 81.27 | 81.17 | 80.01 | 65.56 | 68.76 | 60.63 | 73.18 | - | - |
| MAttNet[28] | vgg16 | Modular-based Models | 80.94 | 79.99 | 82.30 | 63.07 | 65.04 | 61.77 | - | 73.04 | 72.79 |
| DGA[34] | vgg16 | Graph based models | 83.73 | 86.56 | 82.51 | 68.99 | 72.72 | 62.98 | - | 75.76 | 75.79 |
| CMRIN[36] | vgg16 | Graph based models | 84.02 | 84.51 | 82.59 | 71.46 | 75.38 | 64.74 | - | 76.16 | 76.25 |
| MAttNet[28] | resnet101 | Modular-based Models | 85.65 | 85.26 | 84.57 | 71.01 | 75.13 | 66.17 | - | 78.10 | 78.12 |
| DGA[34] | resnet101 | Graph based models | 86.34 | 86.64 | 84.79 | 73.56 | 78.31 | 68.15 | - | 80.21 | 80.26 |
| CMRIN[36] | resnet101 | Graph based models | 86.99 | 87.63 | 84.73 | 75.52 | 80.93 | 68.99 | - | 80.45 | 80.66 |
| CMN[27] | frcnn-vgg16 | Modular-based Models | - | 75.94 | 79.57 | - | 59.29 | 59.34 | 69.30 | - | - |
| ParallelAttn[30] | frcnn-vgg16 | Attention mechanism | 81.67 | 80.81 | 81.32 | 64.18 | 66.31 | 61.46 | - | 69.47 | - |
| LGRANs[33] | frcnn-vgg16 | Graph based models | 82.0 | 81.2 | 84.0 | 66.6 | 67.6 | 65.5 | - | 75.4 | 74.7 |
| GroundNet[38] | frcnn-vgg16 | External parser | - | - | - | - | - | - | 68.9 | - | - |
| listener[25] | frcnn-resnet101 | Joint embedding | 77.48 | 76.58 | 78.94 | 60.5 | 61.39 | 58.11 | 71.12 | 69.93 | 69.03 |
| **spe**+lis+RL[25] | frcnn-resnet101 | Joint embedding | 78.14 | 76.91 | 80.1 | 61.34 | 63.34 | 58.42 | 72.63 | 71.65 | 71.92 |
| spe+**lis**+RL[25] | frcnn-resnet101 | Joint embedding | 78.36 | 77.97 | 79.86 | 61.33 | 63.1 | 58.19 | 72.02 | 71.32 | 71.72 |
| NMTRee[37] | frcnn-resnet101 | External parser | 85.65 | 85.63 | 85.08 | 72.84 | 75.74 | 67.62 | 78.03 | 78.57 | 78.21 |
| CM-Att-Erase[32] | frcnn-resnet101 | Modular-based Models | 87.47 | 88.12 | 86.32 | 73.74 | 77.58 | 68.85 | - | 80.23 | 80.37 |
| VL-BERT$_{BASE}$[40] | frcnn-resnet101 | Language pretraining | - | - | - | 79.88 | 82.40 | 75.01 | - | - | - |
| VL-BERT$_{LARGE}$[40] | frcnn-resnet101 | Language pretraining | - | - | - | 80.31 | 83.62 | 75.45 | - | - | - |

referred objects with 3 referring expressions on average.

The main difference between RefCOCO and RefCOCO+ is that in RefCOCO+, players were disallowed from using location words, therefore focusing the referring expression to purely appearance-based descriptions. RefCOCO consists of 142,209 refer expressions for 50,000 objects in 19,994 images, and RefCOCO+ has 141,564 expressions for 49,856 objects in 19,992 images [19]. The dataset is split into train, validation, Test A, and Test B. Test A spilt contains multiple people, and Test B contains multiple instance of all other objects.

*c) RefCOCOg:* This dataset was collected on Amazon Mechanical Turk, which are similar to the captions in that they are longer and more complex, often entire sentences rather than phrases. Mao et al. [18] construct a Mechanical Turk task in a non-interactive setting, which requires a group to write a natural language referring expression for the object in the MSCOCO image, and then another group is asked to click the specified object for the given referring expression. If the click overlaps with the correct object, the referring expression is considered valid and is added to the dataset. If not, another reference expression is collected for the object. This expression generation and verification tasks are interactively repeated three times. This dataset contains 104,560 expressions for 54,822 objects in 26,711 images. On average, each object has 1.91 expressions, and each image has 3.91 expressions. The average length of RefCOCO expressions is 3.61, while the average length of RefCOCO+ is 3.53 and the average length of RefCOCOg is 8.43 words. The split is 85,474 and 9,536 expression-referent pairs for training and validation.

*d) Flickr 30k entities:* This dataset [60] is for phrase localization, *i.e.*, given an image and a caption that accurately describes it, predicting a bounding box for a specific entity mention from that sentence. There are totally 513,644 entity or scene mentions in the 158,915 Flickr30k [72] captions (3.2 per caption), and these have been linked into 244,035 co-reference chains (7.7 per image). The box drawing process has yielded 275,775 bounding boxes in the 31,783 images (8.7 per image). The annotation process consists of two main stages: co-reference resolution, or forming co-reference chains that refer to the same entities, and bounding box annotation for the resulting chains. This workflow provides two advantages: first, identifying co-reference mentions helps reduce redundancy and save box-drawing effort; and second, co-reference annotation is intrinsically valuable, e.g., for training cross-caption co-reference model.

*e) GuessWhat?!:* GuessWhat dataset [61] is a cooperative two-player game in which both players see the picture of a rich visual scene with several objects. One player – the oracle – is randomly assigned an object in the scene. This object is not known by the other player – the questioner – whose goal is to locate the object, by asking the oracle a series of yes-no questions. The raw GuessWhat?! dataset is composed of 155,280 dialogues containing 821,889 question/answer pairs on 66,537 unique images and 134,073 unique objects. The answers are respectively 52.2% no, 45.6% yes and 2.2% N/A. On average, there are 5.2 questions per dialogue and 2.3 dialogues per image. The dialogues contain 3,986,192-word tokens in total, making up 11,465 different words with at least one occurrence and 5,444 words with at least 3 occurrences.

TABLE V
COMPARISON WITH STATE-OF-THE-ART METHODS ON REFCOCO, REFCOCO+ AND REFCOCOG WHEN DETECTED OBJECTS ARE USED.

| Method | Visual features | Categories | RefCOCO | | | RefCOCO+ | | | RefCOCOg | | |
|---|---|---|---|---|---|---|---|---|---|---|---|
| | | | val | testA | testB | val | testA | testB | val* | val | test |
| MLE[18] | vgg16 | CNN/LSTM framework | - | 58.32 | 48.48 | - | 46.86 | 34.04 | 40.75 | - | - |
| MMI[18] | vgg16 | CNN/LSTM framework | - | 64.90 | 54.51 | - | 54.03 | 42.81 | 45.85 | - | - |
| Visdif[19] | vgg16 | CNN/LSTM framework | - | 62.50 | 50.80 | - | 50.10 | 37.48 | 41.85 | - | - |
| CMN[27] | vgg16 | Modular-based Models | - | 71.03 | 65.77 | - | 54.32 | 47.76 | 57.47 | - | - |
| Luo[22] | vgg16 | CNN/LSTM framework | - | 67.94 | 55.18 | - | 57.05 | 43.33 | - | 49.07 | - |
| Neg Bag[20] | vgg16 | CNN/LSTM framework | 57.3 | 58.6 | 56.4 | - | - | - | 39.5 | - | 49.5 |
| Attribute[24] | vgg16 | Attention mechanism | - | 72.08 | 57.29 | - | 57.97 | 46.20 | 52.35 | - | - |
| VaruContext[23] | vgg16 | Attention mechanism | - | 73.33 | 67.44 | - | 58.40 | 53.18 | 62.30 | - | - |
| MCN[16] | vgg16 | One stage approaches | 75.98 | 76.97 | 73.09 | 62.80 | 65.24 | 54.26 | - | 62.42 | 62.29 |
| MCN[16] | darknet53 | One stage approaches | 80.08 | 82.29 | 74.98 | 67.16 | 72.86 | 57.31 | - | 66.46 | 66.01 |
| FAOA[14] | darknet53 | One stage approaches | 71.15 | 74.88 | 66.32 | 56.86 | 61.89 | 49.46 | - | 59.44 | 58.90 |
| DGA[34] | resnet101 | Graph based models | - | 78.42 | 65.53 | - | 69.07 | 51.99 | - | - | 63.28 |
| ParallelAttn[30] | frcnn-vgg16 | Attention mechanism | - | 75.31 | 65.52 | - | 61.34 | 50.86 | 58.03 | - | - |
| LGRANs[33] | frcnn-vgg16 | Graph based models | - | 76.6 | 66.4 | - | 64.0 | 53.4 | 62.5 | - | - |
| NMTree[37] | frcnn-vgg16 | External parser | 71.65 | 74.81 | 67.34 | 58.00 | 61.09 | 53.45 | - | 61.01 | 61.46 |
| listenser[25] | frcnn-resnet101 | Joint embedding | - | 71.63 | 61.47 | - | 57.33 | 47.21 | 56.18 | - | - |
| spe+lis+RL[25] | frcnn-resnet101 | Joint embedding | - | 69.15 | 61.96 | - | 55.97 | 46.45 | 57.03 | - | - |
| spe+**lis**+RL[25] | frcnn-resnet101 | Joint embedding | - | 72.65 | 62.69 | - | 58.68 | 48.23 | 58.32 | - | - |
| MAttNet[28] | frcnn-resnet101 | Modular-based Models | 76.40 | 80.43 | 69.28 | 64.93 | 70.26 | 56.00 | - | 66.67 | 67.01 |
| MAttNet[28] | mrcnn-resnet101 | Modular-based Models | 76.65 | 81.14 | 69.99 | 65.33 | 71.62 | 56.02 | - | 66.58 | 67.27 |
| NMTRee[37] | frcnn-resnet101 | External parser | 76.41 | 81.21 | 70.09 | 66.46 | 72.02 | 57.52 | 64.62 | 65.87 | 66.44 |
| CM-Att-Erase[32] | frcnn-resnet101 | Modular-based Models | 78.35 | 83.14 | 71.32 | 68.09 | 73.65 | 58.03 | - | 67.99 | 68.67 |
| ViLBERT[39] | frcnn-resnet101 | Language pretraining | - | - | - | 72.34 | 78.52 | 62.61 | - | - | - |
| VL-BERT$_{BASE}$[40] | frcnn-resnet101 | Language pretraining | - | - | - | 71.60 | 77.72 | 60.99 | - | - | - |
| VL-BERT$_{LARGE}$[40] | frcnn-resnet101 | Language pretraining | - | - | - | 72.59 | 78.57 | 62.30 | - | - | - |
| RCCF [17] | DLA-34 | One stage approaches | - | 81.06 | 71.85 | - | 70.35 | 56.32 | - | - | 65.73 |

TABLE VI
RESULTS COMPARISON ON FLICKR30K ENTITIES. PERFORMANCE OF OUR METHOD COMPARED WITH CANONICAL CORRELATION ANALYSIS (CCA) BASELINE ON 100 EDGEBOX PROPOSALS IN FLICKR30K ENTITIES DATASET.

| Model | Categories | Accuracy(%) |
|---|---|---|
| SCRC[21] | CNN/LSTM framework | 27.8 |
| GroundR[41] | Weakly Supervised | 47.81 |
| MCB[26] | Joint embedding | 48.69 |
| ZSGNet(cls)(vgg16)[15] | One stage approaches | 60.12 |
| ZSGNet(cls)(resnet50)[15] | One stage approaches | 63.39 |
| FAOA(Bert)[14] | One stage approaches | 68.69 |
| DDPN(vgg16)[67] | Joint embedding | 70.0 |
| DDPN(resnet101)[67] | Joint embedding | 73.3 |
| CMCC[35] | Graph-based models | 76.74 |

TABLE VII
RESULTS COMPARISON ON THE COPS-REF DATASET.

| Model | Categories | Full | DiffCat | Cat &attr | Cat &cat | Cat | WithoutDist |
|---|---|---|---|---|---|---|---|
| GroundR[41] | Weakly Supervised | 19.1 | 60.2 | 38.5 | 35.7 | 38.9 | 75.7 |
| MattNet[28] | Modular-based | 26.3 | 69.1 | 45.2 | 42.5 | 45.8 | 77.9 |
| CM-Att-Erase[32] | Modular-based | 28.0 | 71.3 | 47.1 | 43.4 | 48.4 | 80.4 |

*f) CLEVR-Ref+:* Above datasets are collected for REC problem in complex, real-world images. However, there have been evidences show that those datasets are biased *i.e.*, simply selecting the salient foreground object will yield a much higher baseline than random guessing. This casts doubts on the true capability of current REC models. Moreover, there are no annotations for the intermediate reasoning process in those datasets, thus their performance evaluation can only depend on the final prediction (*e.g.*, bounding boxes) without considering the intermediate reasoning process of the REC models.

CLEVR-Ref+ [62] considers a different situation, where they use synthesized images where objects are placed on a 2D plane and only have a small number of choices in terms of shape, color, size, and material. The referring expressions are also synthesized using carefully designed templates, which can be very complex and requires a strong reasoning ability. Together with a uniform sampling strategy, this design can mitigate dataset bias and reveal the model's reasoning ability. CLEVR-Ref+ contains 70K images in train set, 15K images in validation and test set, and every image is associated with 10 referring expressions.

*g) Cops-Ref:* In some currently popular referring expressions dataset, the data typically describe only some simple distinctive properties of the object and the images contain limited distracting information, which fail to provide an ideal test bed for evaluating the reasoning ability of the models. To alleviate these problems, Cops-Ref [63] adopts a novel expression engine rendering various reasoning logic that can be flexibly combined with rich visual properties to generate expressions with varying compositionality. Moreover, a new test setting is proposed to better exploit the full reasoning chain embodied in an expression. There are totally 148,712 expressions and 1,307,885 regions on 75,299 images. The average length of the expressions is 14.4 and the size of the vocabulary is 1,596. There are 119,603/16,524/12,586 expressions for training/validation/testing, respectively.

*h) Embodied Referring Expression datasets:* Except for referring expression datasets on static images, there are some datasets move the task to embodied environments.

**Touchdown-SDR** [64] is a dataset for spatial reasoning using real-life visual observations. The agent must first follow the navigation instructions in the real visual city environment, then identify a location described in natural language, and find a hidden object in the target location. Unlike the normal task of referring expression, the goal of the Touchdown-SDR is to describe specific locations rather than distinguish between them.

In the **Refer360°** dataset [65], instructions are given from the perspective of the local field of view (FoV) of the scene, which can be dynamically changed to explore the scene. An important feature of this dataset is that the target location is not an object but any point in the scenario. The Refer360° consists of 17,137 instruction sequences (8.57 per scene) describing randomly distributed target locations from 2,000 panoramic scenarios in the SUN360 dataset.

**REVERIE** [66] combines visual and language navigation and referring expression recognition into one task. In a photo-realistic 3D environment, the system is asked to navigate to the goal location and point out the remote object in the real indoor environment. This dataset consists of 10,318 panoramic images from 86 buildings, containing 4,140 target objects and 21,702 instructions.

*i) Evaluation Measures:* The evaluation is performed by calculating the Intersection over Union (IoU) ratio between the true bounding box and the top predicted box for a referring expression. If the IoU is larger than 0.5, the prediction is considered a true positive. Otherwise, we count it as a false positive. The scores are then averaged over all referring expressions.

*j) Results of existing methods:* We summarize results on the above datasets in Tables II-VII. Firstly, from the results in Table II and III, we observe that supervised methods perform significantly better than weakly supervised methods on ReferItGame dataset. Similar conclusions can also be drawn from the results on Flickr 30k entities dataset (in Table VI) and Cops-Ref dataset (in Table VII). However, weakly supervised methods may be more practical since large-scale fully annotated referring expression datasets are costly.

For supervised methods, we can see that on the most popular dataset, RefCOCO & RefCOCO+, pretraining-based approaches [39], [40] consistently yield the best performance and surpass other approaches by a significant margin. This shows that pretraining on a large-scale vision-and-language dataset can indeed improve the generalization ability of the learned joint vision-and-language representations. Others, like attention-based approaches [30], [31], graph-based approaches [34], [36], or modular-based approaches [28], [32], perform clearly better than most early attempts that follow the CNN/LSTM or joint embedding framework, verifying the effectiveness of the attention mechanism (typically, graph attention, modular attention) on mining the feature relations of multi-modal information. Moreover, compare to the results in Table IV and Table V, we observe that the performance of those methods on human-labeled object proposals are significantly higher than the performance on detected object proposals. This shows that the noise information introduced by the detected object proposals severely hampers the performance. Similar conclusions can be drawn from the results on other datasets.

On the other hand, some one-stage approaches [16], [17], though being much more faster, still perform competitively to many complex attention-based or modular-based two-stage methods, thanks to the use of advanced backbone models, carefully designed learning schemes and label assignment strategies, and maybe also the use of additional annotations [16].

## IV. DISCUSSION AND FUTURE DIRECTIONS

The introduction of Referring expression comprehension (REC) task has aroused great interest and it is initially encouraged by its maturity in the basic tasks of computer vision and natural language processing. Challenges still exists since it requires joint reasoning over the visual and textual information. In this section, we will discuss the limitations of the REC technique and future directions.

### A. What are current limitations of existing REC techniques?

At first, current approaches of REC lack interpretability. Most of the methods we reviewed in this survey are belong to the two-stage category, where a large number of studies have been focused on joint embedding approach. This joint embedding model implicitly fuses expression and candidate objects and then selects the best matching object. Therefore, the process of matching textual and visual contents is more like a black-box in which the reasoning process cannot be visualized. Although the attention mechanism is used, it only shows the attended regions which are more important to the decision process. It is still a challenge to visualise and explain the reasoning steps during the referring expression comprehension process.

In addition, in the process of evaluation, the system can only be conducted on the final prediction and does not evaluate the intermediate step-by-step reasoning process, which is not conducive to the development of interpretable models. Roughly speaking, many models tend to give a direct answer without an intermediate reasoning process. As a result, it is difficult for people to evaluate the reasoning's capability of model and analyze the defects of the model.

Second, current state-of-the-art models suffered serious dataset bias problem. To be specific, there is an imbalance of samples in the current dataset. During the data selection and annotations process, many expressions describe the referents directly with attributes. This imbalance makes the model can only learn the shallow correlation, but not realize the joint understanding of image and text. As stated by Cirik et al. [73], the REC model may locate a target in an image without reasoning, relying solely on memorizing the target through some implicit clues, such as shallow correlations introduced by unexpected biases. In addition, REC models may ignore the expression and also can outperform random guess only using image information. These models fail to truly understand visual and linguistic information and are not applicable in real scenarios. This dataset bias problem is common in vision and language tasks. For example, in the task of Visual Question Answering, models rely heavily on superficial correlations between question and answer, and predict the answer just according to the textual statistical correlations without understanding the visual contents.

*B. Have we reached the upper bound of performance?*

Current REC approaches are more dependent on the recognition of attributes and relationships of objects. While the fact is that existing datasets contain a limited number of attributes. Therefore, REC techniques have reached good performance with these limited datasets. That is to say, designing more sophisticated models does not yield more performance gains, nor does it require more complex reasoning processes on limited datasets. The graph-based models are examples. They did not observe significant improvements on existing referring expression datasets, although more comprehensive relationship encodings are introduced.

*C. What is likely to be the next big leap of techniques in this task?*

Graph-based and modular-based methods achieve the interpretability of reasoning to a certain extent, and language pre-training method establish the corresponding relationship between vision and text concept. Future directions will focus on how to combine language pre-training training with interpretability in REC tasks.

We believe there is still potential to better exploit the concept of Reasoning Chain to solve challenges in referring expression. Reasoning Chain means the logical reasoning steps during the problem solving process. Just as people in the process of reasoning will follow a certain order of rules, reasoning chain aims to consider the referring expression comprehension task from both spatial and semantic perspective. For example, for the expression "Man to the right of the green bag", a reasonable reasoning process should be finding all bags in the image, identifying the green one, and then locating the man to its right. Some datasets required the REC model to reason has been proposed. Liu et al. propose CLEVR-Ref+ [62], which contains rendered images and automatically generated expression, and the objects in images are simple 3D shapes with attributes. As the Figure 10 shows, the REC system needs only two steps to locate the target object in the RefCOCO [59] dataset. While on the CLEVR-Ref+ [62] dataset, the system requires multiple steps of reasoning.

Recently, Chen et al. propose Cops-Ref [63], which covers a variety of reasoning logics that can be combined flexibly and has rich visual properties. They evaluate state-of-the-art REC models on this new dataset and observed a significant drop, compared to their superior performance on conventional referring expression dataset. This suggests most of the models may only learn to fit the bias of the dataset but not learn to reason. The REC system still asks a long and semantic-rich expression to help evaluate the reasoning ability of the models. Developing REC models with 'true' reasoning ability will be the next challenge.

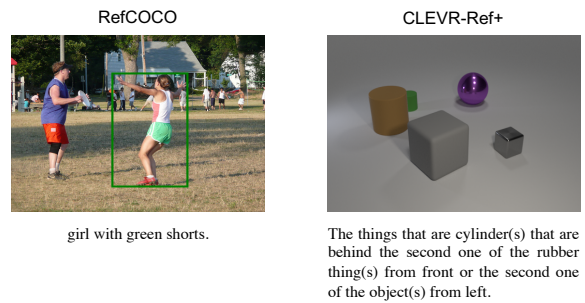

| RefCOCO | CLEVR-Ref+ |
|---|---|
| girl with green shorts. | The things that are cylinder(s) that are behind the second one of the rubber thing(s) from front or the second one of the object(s) from left. |

Fig. 10. Examples from RefCOCO [59] and CLEVR-Ref+ [62] dataset.

## V. CONCLUSION

In this paper we presented a comprehensive review of the state-of-the-art on Referring Expression Comprehension. We first reviewed the most popular approach that embeds referent descriptions and image regions to a common feature space. We then described more advanced modular-based and graph-based models build upon it. Other most recently methods such as one-stage and pre-training models are extensively discussed. After reviewed and compared different referring expression datasets, we then group results according to the datasets, backbone models, settings so that they can be fairly compared and followed. To the best of our knowledge, this is the first comprehensive referring expression comprehension survey paper. Finally, we also pinpointed a number of promising directions for future research. In particular, we suggest to consider complex compositional reasoning in the referring expression so it can benefit other vision-and-language tasks.

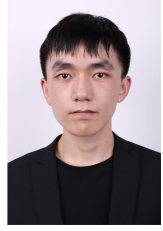

**Chaorui Deng** is a first-year Ph.D student at the University of Adelaide, Australia. His research interests include Image Understanding, Video Understanding, Vision-Language modeling etc.

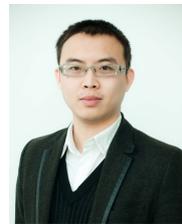

**Qi Wu** is a Senior Lecturer (Assistant Professor) in the University of Adelaide and he is an Associate Investigator in the Australia Centre for Robotic Vision (ACRV). He is the ARC Discovery Early Career Researcher Award (DECRA) Fellow between 2019-2021. He obtained his PhD degree in 2015 and MSc degree in 2011, in Computer Science from the University of Bath, United Kingdom. His educational background is primarily in computer science and mathematics. He works on the Vision and Language problems, including Image Captioning, Visual Question Answering, Visual Dialog etc. His work has been published in prestigious journals and conferences such as TPAMI, CVPR, ICCV, AAAI and ECCV.

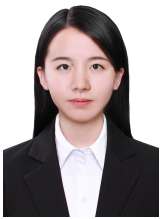

**Yanyuan Qiao** is a first-year Ph.D student at the University of Adelaide, Australia. She obtained her Master degree in 2019, in Electronics and Communications Engineering from the University of Chinese Academy of Sciences, Beijing. And she received the Bachelor degree in 2016, in Sensor Network Technology from the Southeast University, Nanjing, China. She works on the Vision and Language problems, particularly in the area of General Vision and Language Methods in Real Applications.